\documentclass[conference]{IEEEtran}
\usepackage{url}
\usepackage{subcaption}
\usepackage{graphicx}
\usepackage{cite}
\usepackage{amsmath,amssymb,amsfonts}
\usepackage{algorithmic}
\usepackage{textcomp}
\usepackage{xcolor}
\usepackage[hidelinks]{hyperref}       % hyperlinks
\usepackage{booktabs}                  % professional-quality tables
\usepackage{nicefrac}                  % compact symbols for 1/2, etc.
\usepackage{microtype}                 % microtypography
\usepackage{xcolor}                    % colors

\def\BibTeX{{\rm B\kern-.05em{\sc i\kern-.025em b}\kern-.08em
    T\kern-.1667em\lower.7ex\hbox{E}\kern-.125emX}}

\begin{document}

\title{Path of Destruction:\\Learning an Iterative Level Generator Using a Small Dataset}

% The \author macro works with any number of authors. There are two commands
% used to separate the names and addresses of multiple authors: \And and \AND.
%
% Using \And between authors leaves it to LaTeX to determine where to break the
% lines. Using \AND forces a line break at that point. So, if LaTeX puts 3 of 4
% authors names on the first line, and the last on the second line, try using
% \AND instead of \And before the third author name.

\author{\IEEEauthorblockN{Matthew Siper}
\IEEEauthorblockA{\textit{Game Innovation Lab} \\
\textit{New York University}\\
New York, USA \\
ms12010@nyu.edu}
\and
\IEEEauthorblockN{Ahmed Khalifa}
\IEEEauthorblockA{\textit{Institute of Digital Games} \\
\textit{University of Malta}\\
Msida, Malta \\
ahmed@akhalifa.com}
\and
\IEEEauthorblockN{Julian Togelius}
\IEEEauthorblockA{\textit{Game Innovation Lab} \\
\textit{New York University}\\
New York, USA \\
julian@togelius.com}
}

\maketitle

\begin{abstract}
  We propose a new procedural content generation method which learns iterative level generators from a dataset of existing levels. The Path of Destruction method, as we call it, views level generation as repair; levels are created by iteratively repairing from a random starting level. The first step is to generate an artificial dataset from the original set of levels by introducing many different sequences of mutations to existing levels. In the generated dataset, features are observations of destroyed levels and targets are the specific actions that repair the mutated tile in the middle of the observations. Using this dataset, a convolutional network is trained to map from observations to their respective appropriate repair actions. The trained network is then used to iteratively produce levels from random starting maps. We demonstrate this method by applying it to generate unique and playable tile-based levels for several 2D games (Zelda, Danger Dave, and Sokoban) and vary key hyperparameters.
\end{abstract}


\section{Introduction}
Procedural Content Generation (PCG) in games refers to algorithmic methods for generating game content such as quests, items, and levels. Such methods need to generate content that fulfill multiple criteria, both aesthetic and functional. In this paper, we address the question of how can we learn an \emph{iterative} game level generator \emph{from a dataset} of existing levels. Unpacking this statement requires to consider two recent developments in PCG research: PCG via machine learning (PCGML) and PCG via reinforcement learning (PCGRL).

PCGML refers to the use of machine learning, mainly through self-supervised learning on existing content, to learn content generators~\cite{summerville2018procedural}. This provides an alternative to building content generators from scratch, either through ad-hoc methods or using search, optimization, or constraint satisfaction as a principle~\cite{shaker2016constructive, togelius2011search, smith2011answer}. Content generators trained via self-supervised learning tend to produce a whole artifact (such as a level) at the same time; in particular, this is true for deep learning methods such as autoencoders~\cite{jain2016autoencoders} or generative adversarial networks~\cite{volz2018evolving}. In cases where a level has a natural sequential encoding, a sequence learning architecture such as n-grams~\cite{dahlskog2014linear} or LSTM networks~\cite{summerville2016super} can be used, but such encodings are not universally applicable.

PCGRL provides a different way of using machine learning to learn content generation~\cite{khalifa2020pcgrl, earle2021learning}. In the absence of training data, reinforcement learning is used to learn to generate by trial and error. The generator is seen as an agent that takes actions to change its environment, and at the end of an episode it gets rewarded based on the quality of the artifact created by its actions. Interestingly, the resulting generator is fundamentally different from those learned by self-supervised learning, as it produces content \emph{iteratively}, action by action. Iterative content generators have different affordances than one-shot generators. In particular, it can be easier to build mixed-initiative systems around iterative generators~\cite{delarosa2021mixed}. However, reinforcement learning can be slow and unreliable, and designing appropriate rewards is a non-trivial task.

In cases where some prior content already exists, how could we learn an iterative content generator? This paper proposes one of several possible solutions to this problem and applies it to the generation of 2D game levels. Our approach is to artificially destroy levels, framing content generation as the problem of repairing these destroyed levels by learning a repair agent trained from the destruction process.

The first step in our approach is to destroy the levels, which is accomplished by iteratively making destructive edits on the goal map until it is transformed into a randomly generated starting map. This sequence of destructive edits forms the Path of Destruction (PoD), which is then used to form a dataset of desirable repair actions. In this dataset, each instance consists of a feature set, which is a part of the level \emph{after} destruction, and a target, which is a part of the level \emph{before} destruction. Using standard supervised learning methods, we can train a classifier to act as a repair agent by reversing the Path of Destruction. By judiciously choosing the feature set--i.e., the observation space of the agent--and other parameters, we can achieve a high degree of generalization, such that the trained agent not only repairs broken levels but also generates new playable levels from unseen randomized starting levels.

\section{The Path of Destruction}\label{sec:pod_method}
We introduce a technique which we refer to as the Path of Destruction (PoD) in order to train generators to produce unique and playable levels~\footnote{\url{https://github.com/matt-quant-heads-io/path_of_destruction}}\footnote{The name comes from the fact that the dataset is formed from a trace of destructive events (random tile changes), which is then retraced.}. As previously explained, PoD is a technique that consists of generating a training dataset by iteratively destroying goal levels until they reach random noisy levels. During each of these destructive sequences, a repair sequence is iteratively produced with each destructive step and added to the training trajectory, which is used to train the repair agent.

\subsection{Training set generation}

\begin{figure}[ht]
  \centering
  \includegraphics[width=0.95\columnwidth]{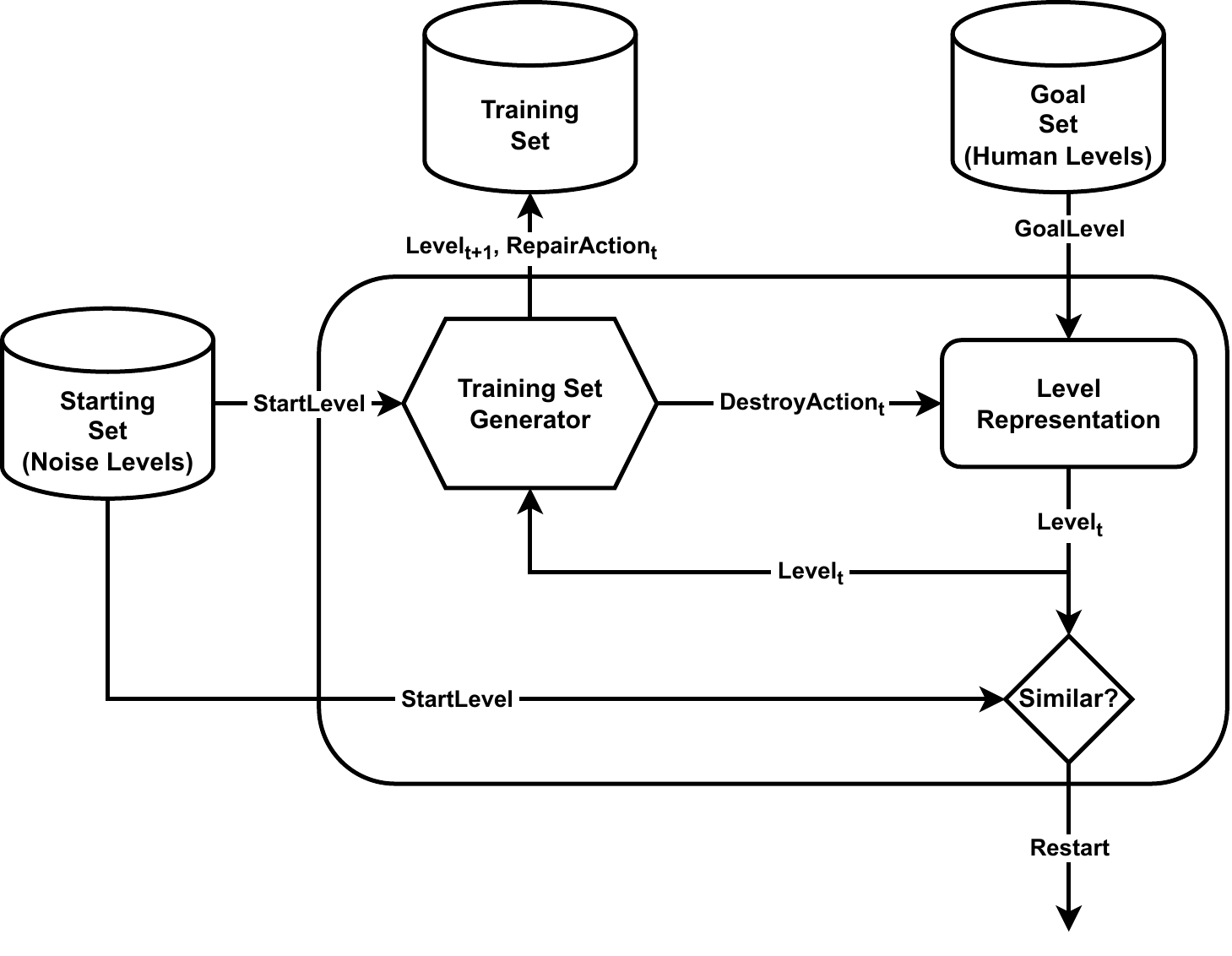}
  \caption{System diagram explaining the training set generation step in the Path of Destruction method.}
  \label{fig:training_set}
\end{figure}

The steps for generating the training set is at the heart of the Path of Destruction technique. The goal of this step is to generate a large artificial dataset that can be used to train an agent to repair destroyed levels and create new levels by ``repairing'' from noise. Figure~\ref{fig:training_set} shows the system diagram that illustrates the main loop of generating the training set. To start this step, we need two datasets:
\begin{itemize}
    \item \textbf{Starting set:} This is a set of ``levels'' that will be used later as the starting level for generation. In this project, the starting set consist of noise (random tiles) that is produced from a predefined distribution such as a uniform distribution.
    \item \textbf{Goal set:} This is a set of levels that we want our generator to learn to repair towards them. In this project, the goal set is a set of human designed levels that are playable and unique.
\end{itemize}

From this dataset, we generate the training dataset by executing the following algorithm:
\begin{enumerate}
    \item Generate a noisy random level from a predefined distribution ($StartLevel$).
    \item Select the closest goal level ($GoalLevel$) from the goal set to the start level ($StartLevel$).
    \begin{enumerate}
        \item Calculate the hamming distance between each level in the goal set and the start level ($StartLevel$).
        \item Select the level with the smallest hamming distance to be the goal level ($GoalLevel$)
    \end{enumerate}
    \item Set the current level ($Level_t$) to be equal to the goal level ($GoalLevel$).
    \item Destroy the current level ($Level_t$) until it is equal to the start level ($StartLevel$).\label{list:destruction}
        \begin{enumerate}
            \item Select the next tile location on the level (can be random or sequential starting from location (0,0)).~\label{list:destruct_traj}
            \item Save the tile value at the selected location from both the current level ($RepairAction_t$) and from the start level ($DestroyAction_t$).
            \item Update the current level ($Level_t$) with the tile value saved from the start level ($DestroyAction_t$).
            \item Add the current level and the original tile value before destruction ($Level_t$, $RepairAction_t$) to the training set.
            \item If the current level ($Level_t$) and the start level ($StartLevel$) are not the same then go back to step~\ref{list:destruct_traj}.
        \end{enumerate}
    \item Repeat the entire process until the training set reaches the appropriate size.
\end{enumerate}

In our experiments, the destruction process (step~\ref{list:destruction}) stops when the current level ($Level_t$) is the same as the start level ($StartLevel$). However, this is not a requirement. The only stopping requirement is that the current level ($Level_t$) is close to the starting set distribution. This allows the network to learn to repair (generate) levels from any sampled starting level that follows that distribution which is explained in section~\ref{sec:generation}. One could imagine the stopping criteria be when the current level ($Level_t$) is \emph{close enough} to the start level ($StartLevel$) instead of identical or the tile distribution of the current level ($Level_t$) is close enough to that of the starting set.

\subsection{Level generator training}~\label{sec:gen_training}
For this project, we train a neural network that takes a cropped observation of the current level as input, and predicts the appropriate repair action. The action space is defined as the set of possible game tile values. The cropped observation consists of the tiles surrounding and including the current tile location. The observation is parameterized by a number, which we refer to as the crop size. The crop size determines the number of tiles surrounding the current tile location that are observable by the agent at each step. The input observation consists of the tiles cropped around the changed tile such that the changed tile is in the center (shown in figure~\ref{fig:observations}). We experimented with different crop sizes as previous research~\cite{ye2020rotation,earle2022illuminating} showed that neural networks with local observation are cable of learning a more general strategy compared to networks that have access to the full observation.

\subsection{Level generation}~\label{sec:generation}

\begin{figure}[ht]
  \centering
  \includegraphics[width=0.75\columnwidth]{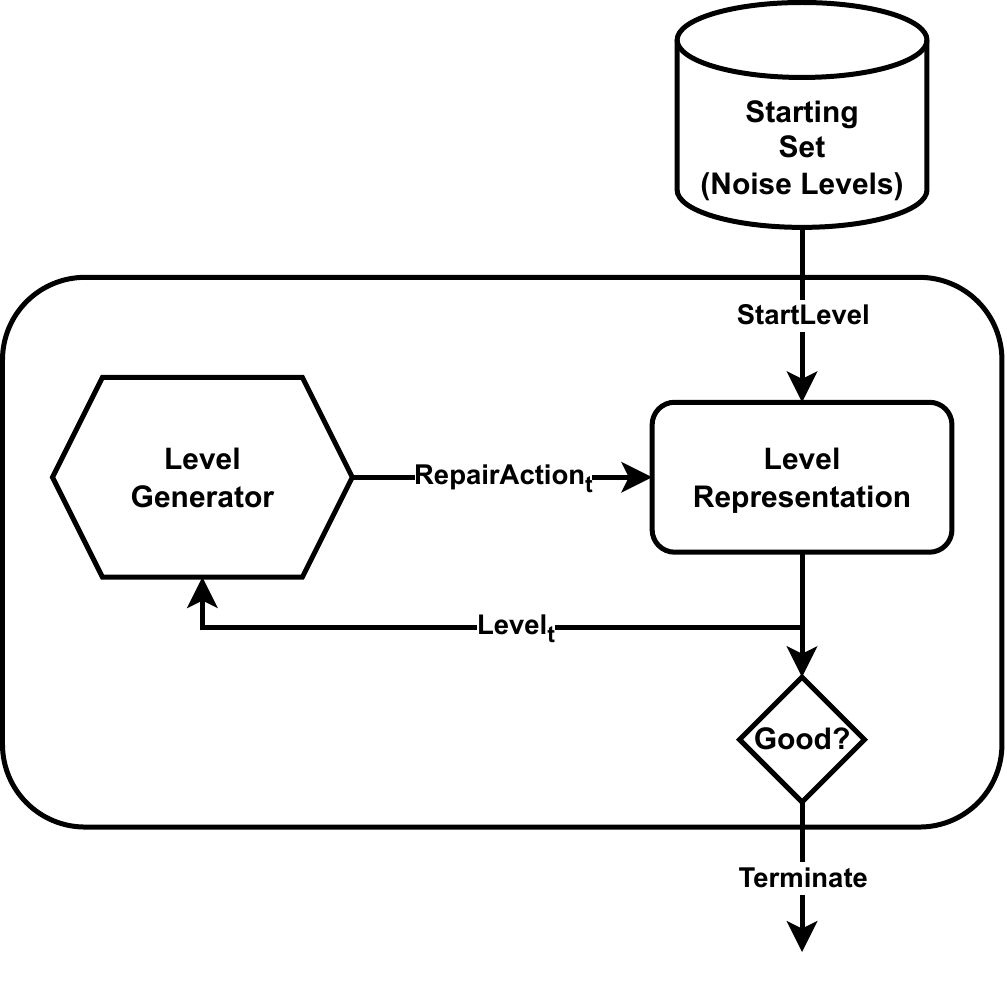}
   \caption{System diagram explaining the level generation step in the Path of Destruction method.}
   \label{fig:generation_step}
\end{figure}

The inference, or level generation step, is the final step of the Path of Destruction. This step focuses on converting a noisy starting level to a playable level. Figure~\ref{fig:generation_step} shows the main loop of generating a new level. The level generator modifies the current level iteratively with following these steps:
\begin{enumerate}
    \item Initialize the current level ($Level_t$) via generating a noisy level from a predefined tile distribution.
    \item Select the next tile location \label{list:generation_start} either randomly or sequentially starting from (0,0) location.
    \item Feed the cropped current level ($Level_t$) to network, which outputs a repair action ($RepairAction_t$).
    \item Update the current tile location in the current level ($Level_t$) with the repair action ($RepairAction_t$).
    \item If the current level ($Level_t$) isn't playable and the number of steps hasn't reached the threshold, go to step~\ref{list:generation_start}.
\end{enumerate}
This process enables the agent to transform a starting noisy level into a playable level ($Level_t$) over sequential iterations of repair. The stopping condition for the generation process is usually defined by the user based on the current application. In our project, we used playability as the stopping condition since we are trying to generate playable levels.

\section{Experiments}
We conduct 4 different experiments. In the first experiment, we measure the effect of constraining the observation size. In the second experiment, we constrain the goal set size to measure its impact on agent performance. In the third experiment, we evaluate the robustness of the PoD technique by applying it to additional game types. Finally, we benchmark our PoD algorithm against a state of the art level generation method that uses Conditional Embedding Self-Attention Generative Adversarial Networks (CESAGAN)~\cite{torrado2020bootstrapping}. For our level generator, we used a convolutional neural network. The network consists of 3 convolution layers, each with $3x3$ kernels of size $128$, $128$, and $256$, respectively. A $2x2$ max pooling layer is used after the second convolution layer, followed by a fully connected layer and a softmax activation function. The network is trained using the Rmsprop algorithm with a batch size of $64$ and a learning rate of $0.001$. Training consists of $500$ epochs and employs categorical cross entropy as the loss function. For every experiment, we trained $3$ networks and calculated the mean and standard deviation to show training stability.

For the first two experiments, and the last experiment, we used the Zelda version from PCGRL as our test bed. The games tested and their playability conditions are described in section~\ref{sec:diff_games}. After we trained the level generators, we evaluated each agent’s performance over $10,000$ independent inference trials. We calculated the percentage of generated playable levels, as well as, the percentage of generated playable and unique levels. For measuring uniqueness, we used the hamming distance between all the playable levels (generated or goal) and removed all of the levels that were less than 10\% different. For the benchmarking experiment, we computed the number of duplicate levels instead of similarity to align with the metrics used in the original CESGAN experiments.

\subsection{Constraining observation size}

\begin{figure}[ht]
    \centering
    \begin{subfigure}[t]{0.23\linewidth}
        \centering
        \raisebox{0.2\height}{\includegraphics[width=\linewidth]{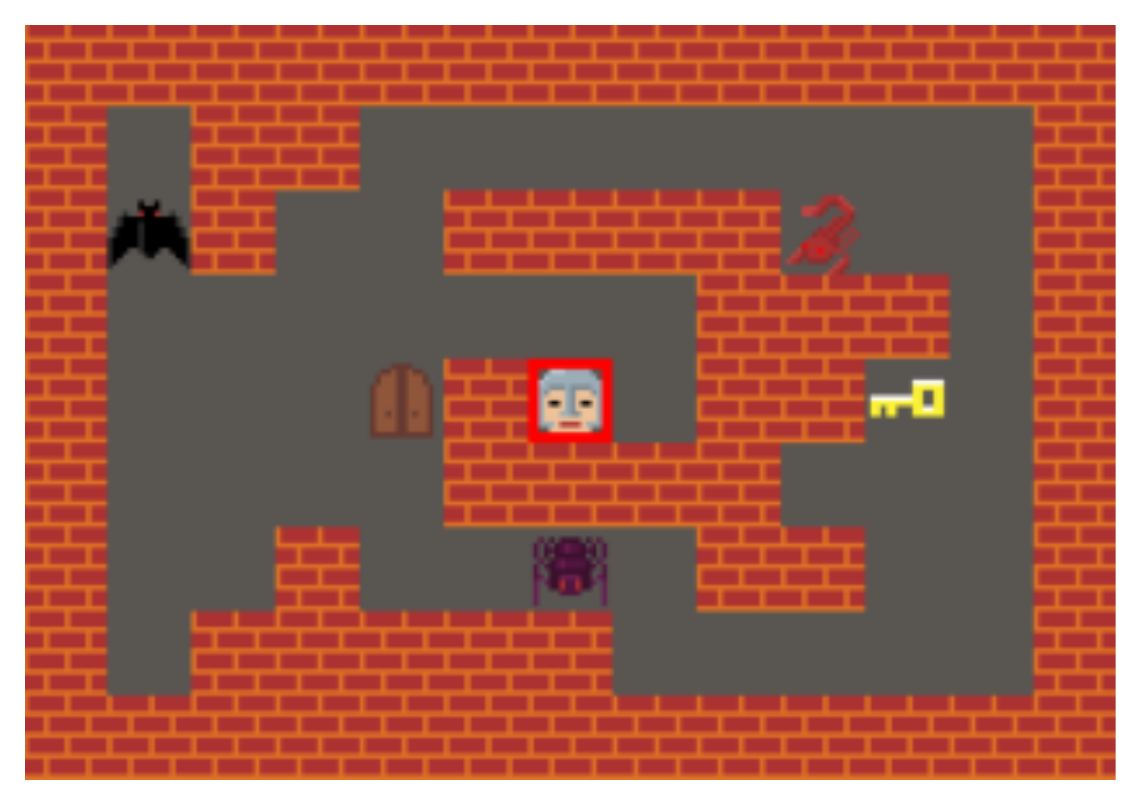}}
        \caption{Actual Map}
        \label{fig:observation_goal}
    \end{subfigure}
    \begin{subfigure}[t]{0.23\linewidth}
        \centering
        \includegraphics[width=\linewidth]{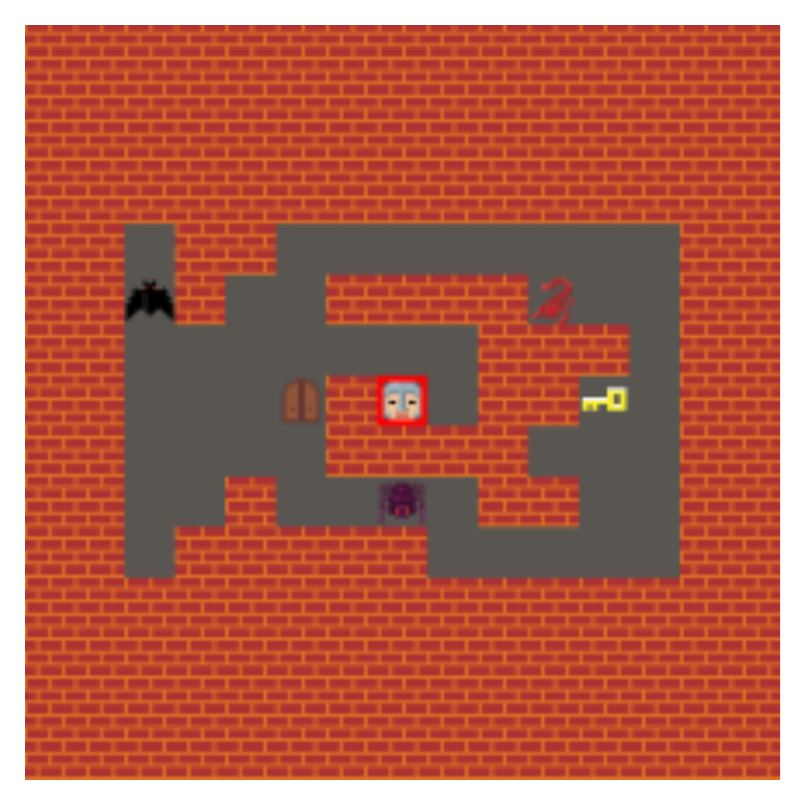}
        \caption{Obs Size 15}
        \label{fig:observation_15}
    \end{subfigure}
    \begin{subfigure}[t]{0.23\linewidth}
        \centering
        \includegraphics[width=\linewidth]{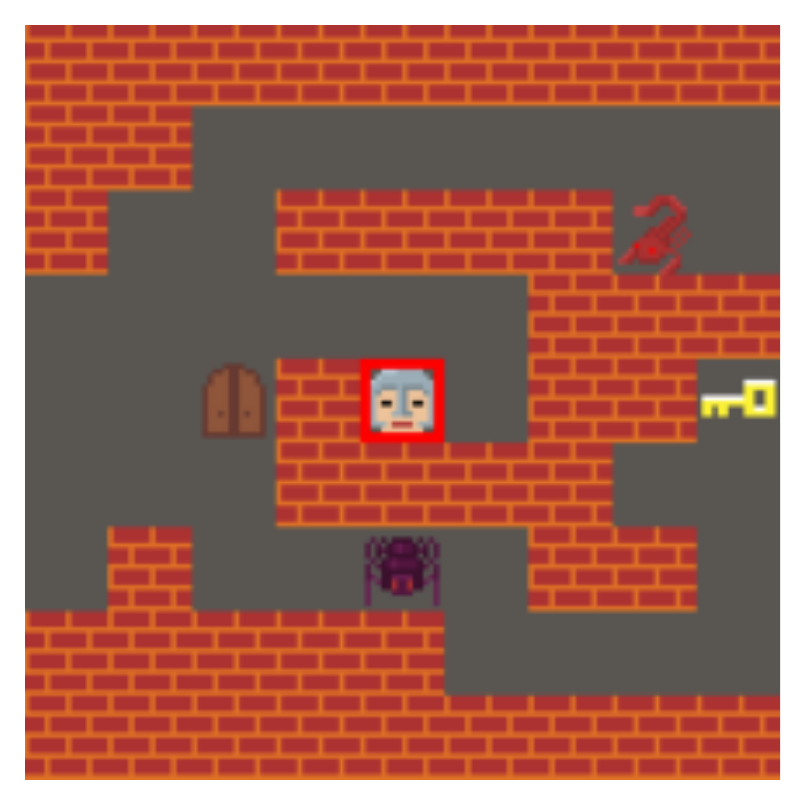}
        \caption{Obs Size 9}
        \label{fig:observation_9}
    \end{subfigure}
    \begin{subfigure}[t]{0.23\linewidth}
        \centering
        \includegraphics[width=\linewidth]{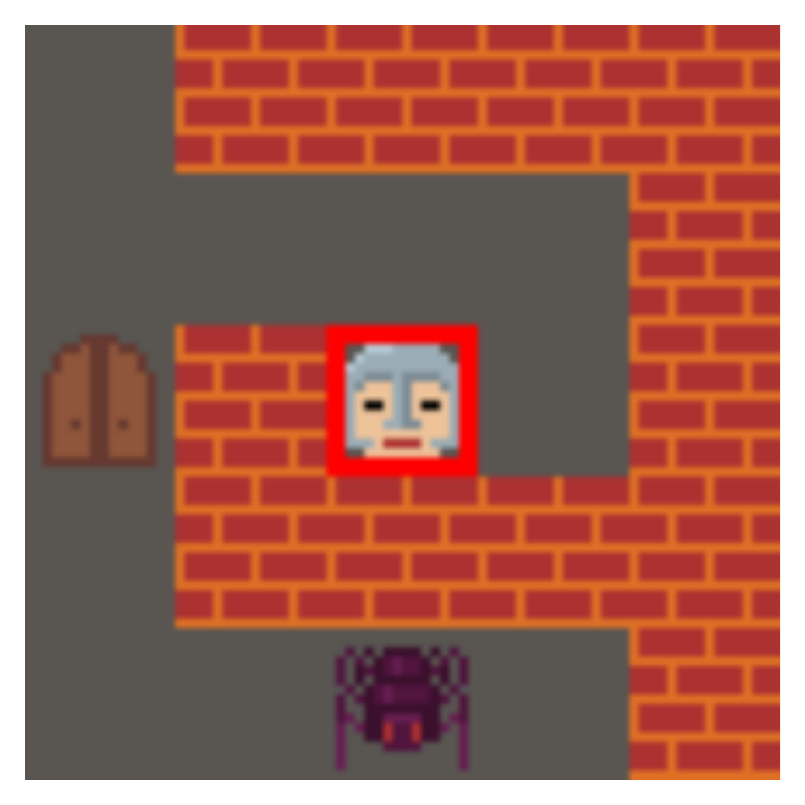}
        \caption{Obs Size 5}
        \label{fig:observation_5}
    \end{subfigure}
    \caption{Example of the different observation sizes for a certain goal map. The red rectangle is the current tile that is going to be acted upon by the agent.}
    \label{fig:observations}
\end{figure}

In this experiment, we explored the effect of crop size in the training set on agent performance. As explained in section~\ref{sec:pod_method}, we converted the destroyed levels into observations by setting the center of the observation around the destroyed tile and cropping the surrounding tiles. This is similar to the narrow representation presented in the PCGRL framework~\cite{khalifa2020pcgrl}. To measure the effect of crop size on agent performance, we tested 3 different observation sizes: $15$, $9$, and $5$. These sizes were selected based on the map size of Zelda ($7 x 11$). A crop size of $15$ gave a full view of the level, whereas a crop size of $9$ gave a partial view, and a crop size of $5$ gave the most local view. Figure~\ref{fig:observation_goal} shows the different observation sizes for the center location of the map. We hypothesized that a smaller observation size would help the trained agent to generate a diverse and unique set of playable levels.

\subsection{Constraining goal set size}
In this experiment, we sought to capture the effect of the size of the goal set used to generate the training data. We tested 3 different goal set sizes: $1$, $5$, and $50$. We expected that a larger goal set size would help the agent produce a more diverse and unique set of playable levels. For each of these experiments, the observation size was fixed to $5$.

\subsection{Different games}\label{sec:diff_games}

\begin{figure}[ht]
    \centering
    \begin{subfigure}{\linewidth}
        \centering
        \includegraphics[width=\linewidth]{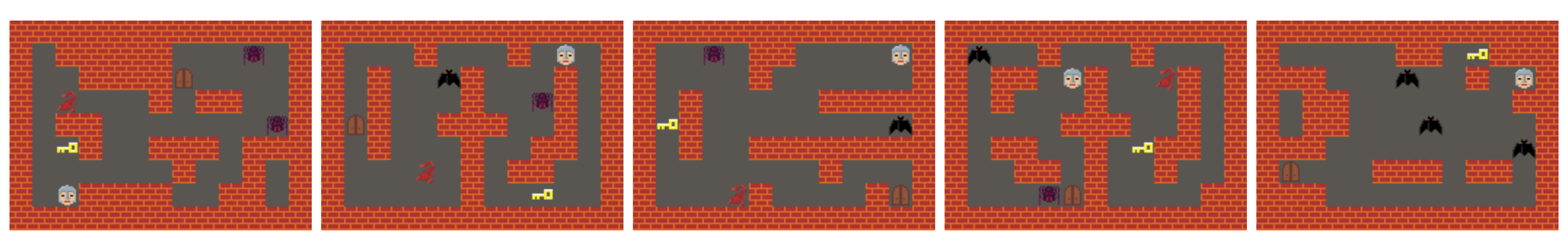}
        \caption{Examples of Zelda levels from the goal set used to generate the training set.}
        \label{fig:goal_zelda}
    \end{subfigure}
    \begin{subfigure}{\linewidth}
        \centering
        \includegraphics[width=\linewidth]{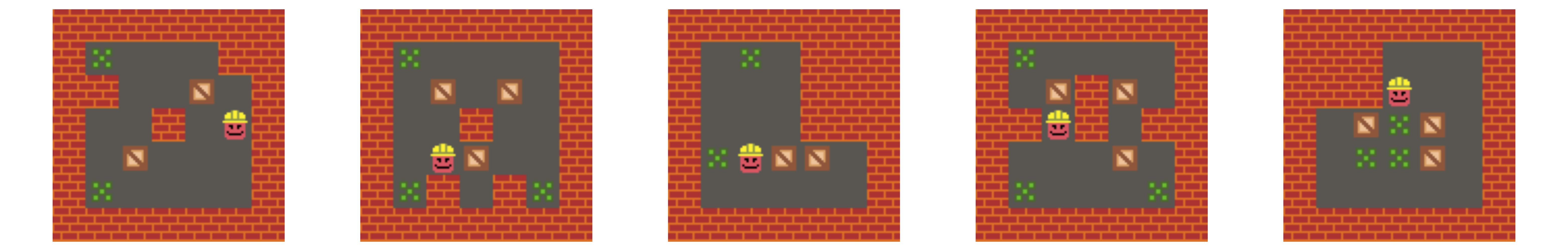}
        \caption{Examples of Sokoban levels from the goal set used to generate the training set.}
        \label{fig:goal_sokoban}
    \end{subfigure}
    \begin{subfigure}{\linewidth}
        \centering
        \includegraphics[width=\linewidth]{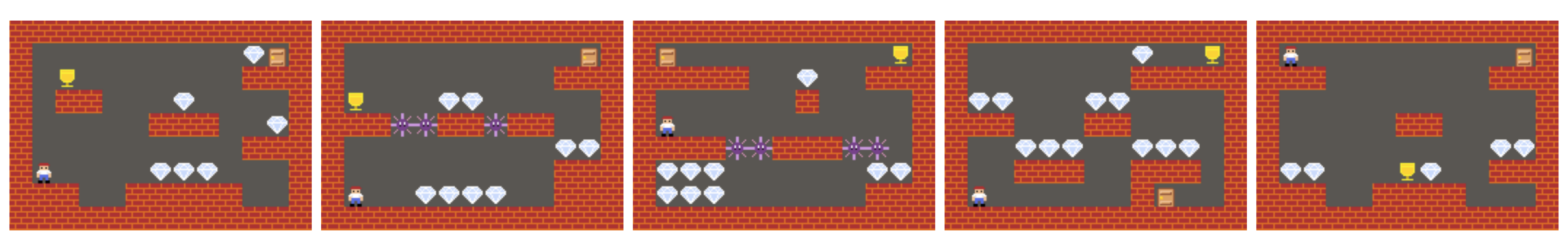}
        \caption{Examples of Danger Dave levels from the goal set used to generate the training set.}
        \label{fig:goal_ddave}
    \end{subfigure}
    \caption{Goal sets for different games used in training}
    \label{fig:goal_set}
\end{figure}

In this experiment, we used the PoD method to train a level generator across 3 different games. Figure~\ref{fig:goal_set} shows the 5 goal levels used for each of the 3 games. The 3 different games that we selected were as follows.
\begin{itemize}
    \item \textbf{Zelda:} is a simple remake of the dungeon system of The Legend of Zelda (Nintendo, 1986). The goal of the game is to get the key and go to the door without dying. The player can either avoid enemies or hit them with their sword for extra points. The goal set levels that we used were all $7x11$ tiles. To measure playability, we ensured that there was one key, one door, and one player. Also, the player must've been able to reach the key and the door.
    \item \textbf{Sokoban:} is a japanese puzzle game (Thinking Rabbit, 1982). The goal of the game is to push the crates to a certain target location without getting stuck. The goal levels that we used were of size $5x5$ tiles. To measure playability, we needed to have one player and a number of crates equal to the number of targets. Lastly, the A* algorithm must've been able to solve the puzzle in a reasonable amount of time.
    \item \textbf{Danger Dave:} is a simple remake of the PC game Dangerous Dave (John Romero, 1988). This game is different than the other two games as it is a platformer game where the player can move left, right, and jump. The goal of the game is to collect the chalice and then reach the door while avoiding the spikes. The player can collect diamonds to get a higher score. The goal levels used were all of size $7x11$ tiles. To measure playability, there must've been one chalice, one door, and one player. Lastly, the A* algorithm must've been able to solve the puzzle in a reasonable amount of time.
\end{itemize}
For all 3 games, we used a goal set size of $5$ because it is a reasonable number of levels that can easily be found in most games. For both Zelda and Danger Dave, we used an observation size of $5$ because both games have the same size map. For Sokoban, we used an observation size of $3$ since the map size is much smaller.

\subsection{Different Algorithms}
In the last experiment, we benchmark our results against the CESAGAN architecture~\cite{torrado2020bootstrapping} for Zelda. The authors of this work introduce a bootstrapping training procedure in which a \textit{Conditional Embedding Self-Attention Generative Adversarial Network} is trained to generate Zelda levels using the same goal set of 50 Zelda maps that we used in our experiments. The network performance is evaluated by computing the percentage of playable levels and the percentage of duplicated levels generated. Table \ref{table:metric_benchmark} shows the percentage of levels generated that are duplicates. We computed these values to align with the metrics used in the CESAGAN work. We compared the CESGAN baseline to the PoD performance, using a goal set size of 50.

\section{Results}
As previously mentioned, for each experiment, we trained 3 networks to ensure stability. Each trained network generated $10,000$ levels, which we used to measure agent performance. In the following subsections, we show these performance results.

\subsection{Observation size}

\begin{table}
  \centering
  \begin{subtable}[t]{0.48\textwidth}
      \centering
      \resizebox{\textwidth}{!}{%
      \begin{tabular}{lll}
        \toprule
        % \multicolumn{2}{c}{Part}                   \\
        \cmidrule(r){1-2}
        Observation size     & Playable     & Playable \& Unique \\
        \midrule
        5        & 37.98 ± 0.18\%  & 37.77 ± 0.1\%     \\
        9        & 53.93 ± 2.82\%  & 28.55 ± 1.25\%      \\
        15       & 86.82 ± 1.34\%  & 18.34 ± 1.0\%  \\
        \bottomrule
      \end{tabular}
      }
      \caption{Agent performance by observation size}
      \label{table:metric_obs}
  \end{subtable}
  \begin{subtable}[t]{0.48\textwidth}
      \centering
      \resizebox{\textwidth}{!}{%
      \begin{tabular}{lll}
        \toprule
        % \multicolumn{2}{c}{Part}                   \\
        \cmidrule(r){1-2}
        Goal size     & Playable     & Playable \& Unique \\
        \midrule
        1        & 83.84 ± 3.57\% & 14.4 ± 0.26\%     \\
        5        & 38.81 ± 1.85\% & 28.47 ± 2.08\%      \\
        50       & 37.98 ± 0.18\% & 37.77 ± 0.1\%  \\
        \bottomrule
      \end{tabular}
      }
      \caption{Agent performance by goal size}
      \label{table:metric_goal}
  \end{subtable}
  \caption{Agent performance by observation size and goal set size}
  \label{table:metrics}
\end{table}

Table~\ref{table:metric_obs} shows the percentage of playable levels and playable and unique levels generated across different observation sizes. It is evident that the size of the observation is positively correlated to the number of playable levels generated, and negatively correlated to the number of playable and unique levels generated. This is likely due, in part, to the fact that a larger observation size makes it easier for the network to understand and correlate what makes a level playable. However, the larger observation size also makes the network prone to overfitting on the training data. This was expressed as a mode collapse (all levels are similar).

\begin{figure}[ht]
    \centering
    \includegraphics[width=\linewidth]{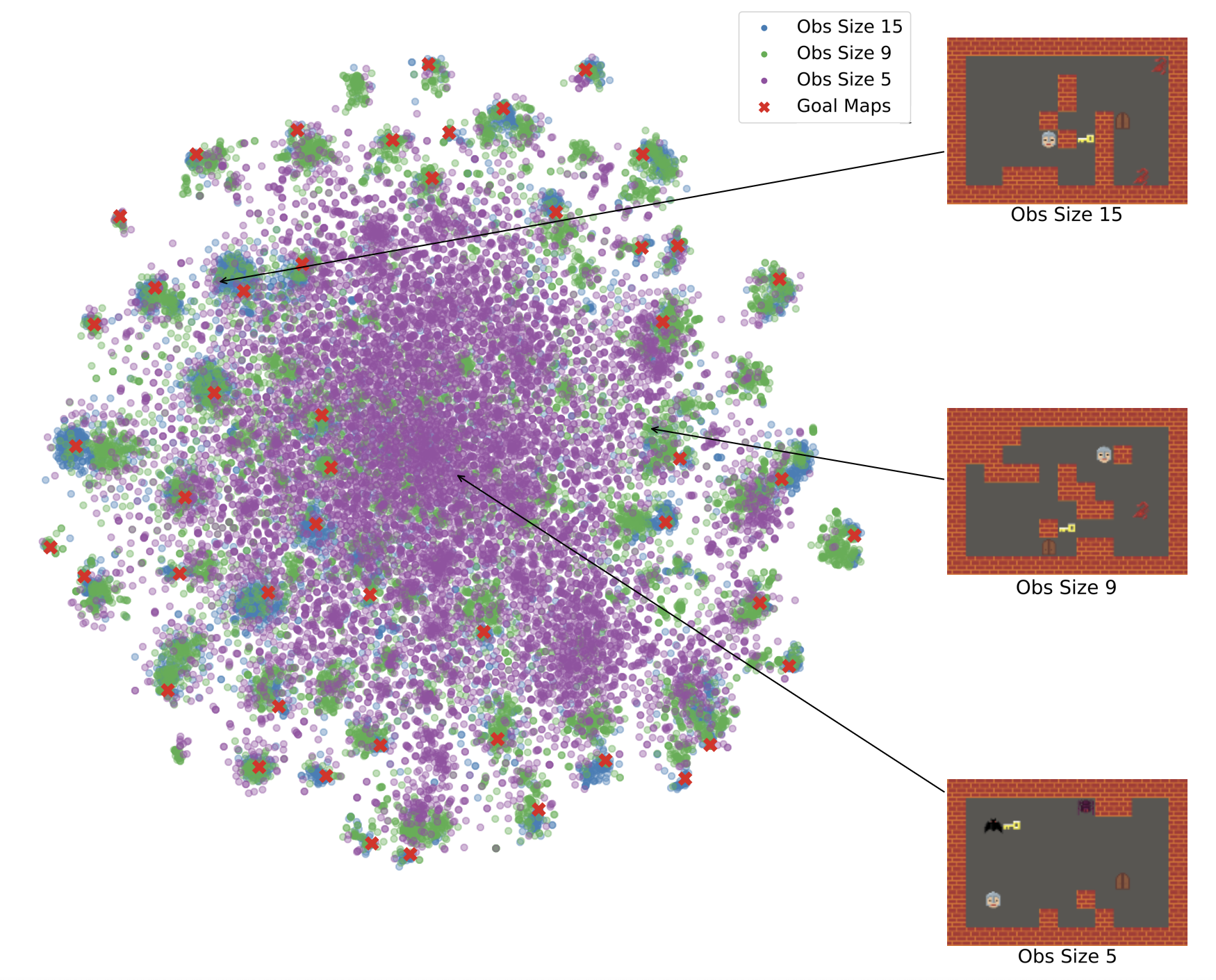}
    \caption{t-SNE visualization of the unique and playable generated levels for the different trained models based on observation size.}
    \label{fig:tsne_obs}
\end{figure}

\begin{figure}[ht]
    \centering
    \includegraphics[width=\linewidth]{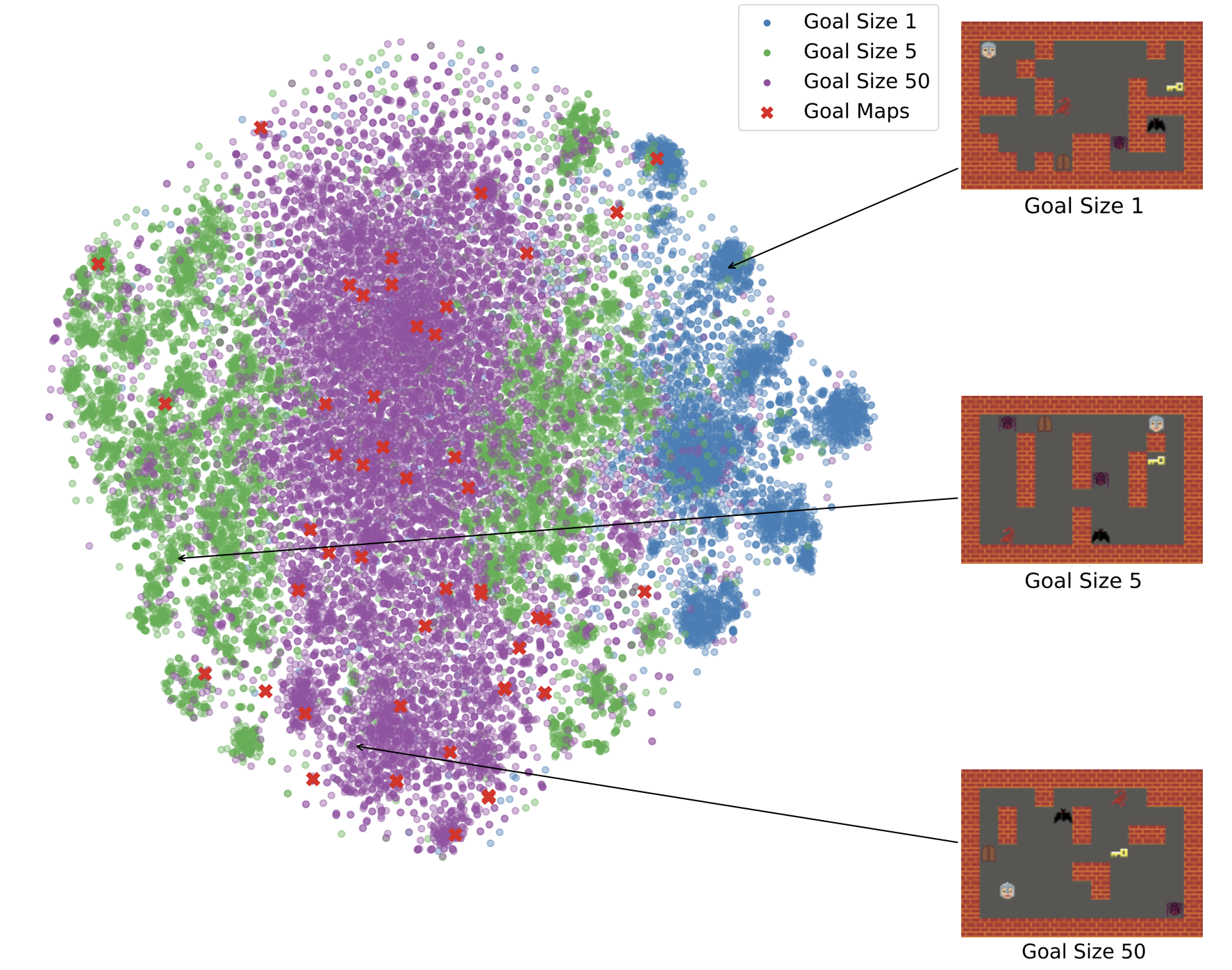}
    \caption{t-SNE visualization of the generated levels for the different trained models based on using different goal set size.}
    \label{fig:tsne_goal}
\end{figure}

Figure~\ref{fig:tsne_obs} uses the t-SNE algorithm to plot the generated unique and playable levels for each observation size. The purpose of this visualization is to show the relation of these levels with respect to each other and the goal maps. It is evident that the generated levels from observation size 15 are mostly clustered around the goal maps.

\subsection{Goal set size}

Table~\ref{table:metric_goal} shows the percentage of playable levels and playable and unique levels generated across different goal set sizes. As is evident in the results, the behavior is similar to what we observed in the observation size experiment. We see that a small goal set size forces the trained generator to overfit on the training data, resulting in generated levels that are less diverse. By contrast, as the goal set size increases, so to does the diversity of the generated levels. Using a larger goal set size to train the network pushes the network to learn a general strategy that doesn't mode collapse.

From the 2D projection of the generated levels shown in figure~\ref{fig:tsne_goal}, we can see that having a larger goal set helps the network to learn to generate levels that are between the goal levels. We noticed that since not all of the 50 levels are very distinct, this could be a limiting factor in the variability of the generated maps. One interesting direction that we did not explore deeply was the mechanism for selecting the goal map in the Path of Destruction method. Since we select the goal maps that are closest to the respective randomly generated starting maps, it is not immediately evident how this selection mechanism affects the diversity of the training set. Furthermore, some goal levels might be more unique than others. Similarly, some goal levels might be more difficult than others. We think the goal level selection method should be investigated in future work.

\subsection{Different games}

\begin{figure}[ht]
    \centering
    \begin{subfigure}{\linewidth}
        \centering
        \includegraphics[width=\linewidth]{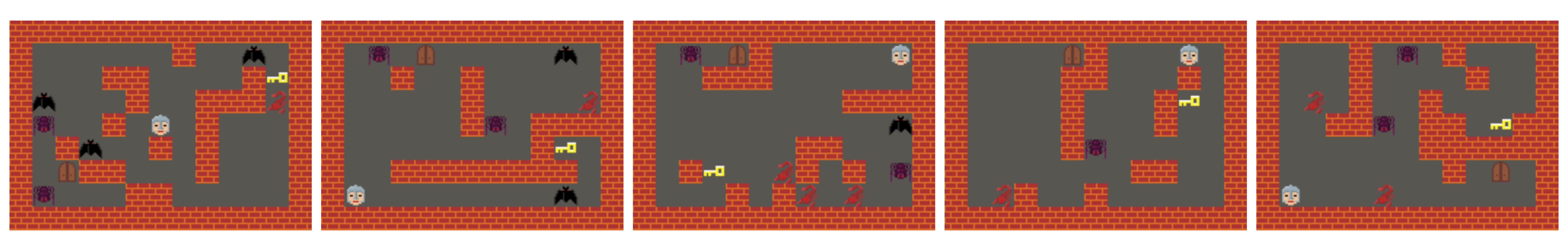}
        \caption{Examples of playable and uniquely generated Zelda levels.}
        \label{fig:examples_zelda}
    \end{subfigure}
    \begin{subfigure}{\linewidth}
        \centering
        \includegraphics[width=\linewidth]{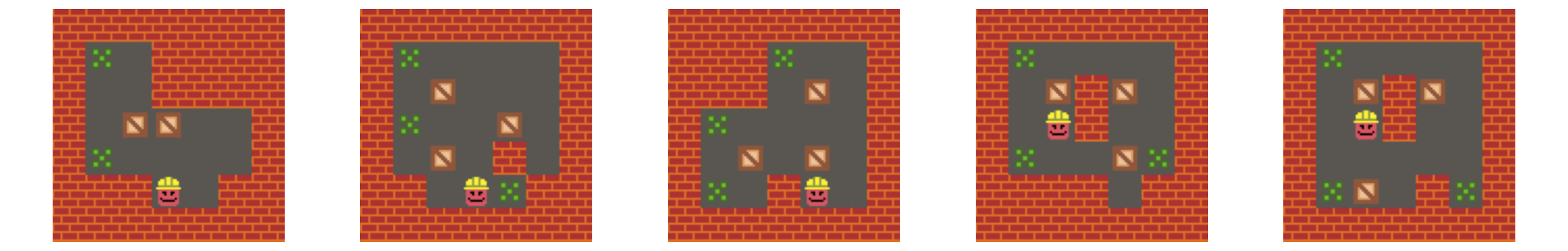}
        \caption{Examples of playable and uniquely generated Sokoban levels.}
        \label{fig:examples_sokoban}
    \end{subfigure}
    \begin{subfigure}{\linewidth}
        \centering
        \includegraphics[width=\linewidth]{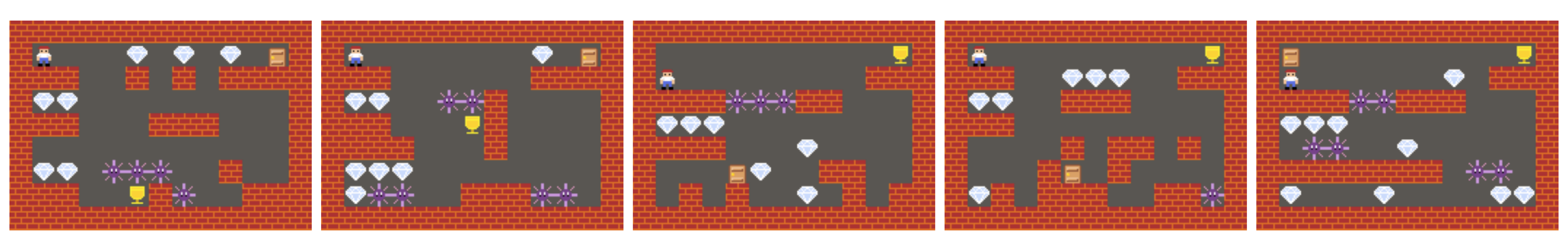}
        \caption{Examples of playable and uniquely generated Danger Dave levels.}
        \label{fig:example_ddave}
    \end{subfigure}
    \caption{Examples of playable and uniquely generated levels for Zelda, Sokoban, and Danger Dave.}
    \label{fig:examples_generated}
\end{figure}

Figure~\ref{fig:examples_generated} shows some of the playable and unique levels generated for each of the three games. The levels are sorted from left to right based on how different they are relative to each other and to the goal levels. In comparing these levels to the 5 trained goal levels shown in figure~\ref{fig:goal_set}, we can see that levels for Sokoban and Danger Dave look more similar to the goal levels than for Zelda. This is also signified in table~\ref{table:metric_games}, where the percentage of playable and unique levels is much less compared to Zelda. We think that the low diversity in Sokoban is due to the smaller level size which makes the 3x3 observation window more global. This is similar to the observation size of 9 used in Zelda. On the other hand, the Danger Dave agent has low scores for both playability and playability and uniqueness because it is a harder game to generate playable levels for. Incorporating the solution path as part of the goal maps, like in the work done using Multidimensional Markov Chains for level generation~\cite{snodgrass2017procedural}, could improve the overall quality in platformer games.

\begin{table}
  \centering
  \begin{tabular}{lll}
    \toprule
    % \multicolumn{2}{c}{Part}                   \\
    \cmidrule(r){1-2}
    Game     & Playable     & Playable \& Unique \\
    \midrule
    Zelda          & 38.81 ± 1.85\% & 28.47 ± 2.08\%     \\
    Sokoban        & 53.31 ± 0.58\%  & 4.81 ± 0.02\%     \\
    Danger Dave    & 18.2 ± 0.35\% & 10.48 ± 0.02\%     \\
    \bottomrule
  \end{tabular}
  \caption{Agent performance across different games}
  \label{table:metric_games}
\end{table}

Although the levels look similar to the goal maps, they have different solutions. This is especially true for Sokoban and Danger Dave because these games are more strategic and thus require more planning compared to Zelda. Since our diversity metric (hamming distance) doesn't take in account the level solution, we think that a lot of interesting Sokoban levels were subsequently removed. For example, if we change just the target location for one of the puzzles, we could then produce a more complex level that requires many more steps to solve even though the hamming distance will be minimally affected. We think that this is an interesting avenue to explore in future work as it could be used to influence the training data generation steps by selecting goal levels that have high diversity more frequently.

\subsection{Different Algorithms}

\begin{table}
  \centering
  \begin{tabular}{lll}
    \toprule
    % \multicolumn{2}{c}{Part}                   \\
    \cmidrule(r){1-2}
    Model     & Playable     & Duplicated levels \\
    \midrule
    CESAGAN            & 47.00\% & 60.30\%     \\
    PoD     & 37.98\% & 0.00\%      \\
    \bottomrule
  \end{tabular}
  \caption{Agent performance compared to baseline}
  \label{table:metric_benchmark}
\end{table}

\begin{figure}[ht]
    \centering
    \begin{subfigure}{\linewidth}
        \centering
        \includegraphics[width=\linewidth]{images/examples/example_zelda_5_5.pdf}
        \caption{Examples of playable and non-duplicated Zelda levels generated by PoD.}
        \label{fig:examples_pod}
    \end{subfigure}
    \begin{subfigure}{\linewidth}
        \centering
        \includegraphics[width=\linewidth]{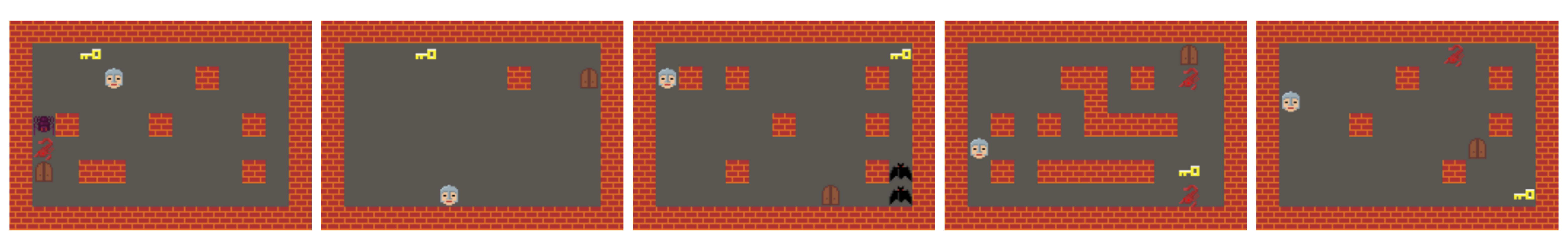}
        \caption{Examples of playable and non-duplicated Zelda levels generated by CESAGAN.}
        \label{fig:examples_gan}
    \end{subfigure}
    \caption{Examples of playable non-duplicated levels generated by Path of Destruction (PoD) and Conditional Embedding Self-Attention Generative Adversarial Network (CESAGAN), respectively.}
    \label{fig:examples_benchmark}
\end{figure}

As is evident from table \ref{table:metric_benchmark}, our PoD agent significantly outperformed the baseline in terms of the percentage of duplicate levels generated. The PoD agent did so while remaining competitive in terms of the percentage of playable levels generated. Figure~\ref{fig:examples_benchmark} shows examples from PoD and CESAGAN, respectively. The levels generated by the PoD have similar structure to the original levels compared to the CESAGAN levels. CESAGAN levels are more open with less connected tiles. We think that the locality of observation forced the PoD network to learn local patterns such as walls needing to be connected to each other. By contrast, the CESAGAN results seemed to learn a general distribution of the positions of wall tiles. We also noticed that levels generated by CESAGAN tended to have fewer enemies compared to PoD levels. The small number of enemies is closer to the actual number of enemies in the goal levels. We believe that locality is the main driver of this. Having a local view prevents the network from easily learning how to count the number of enemies in the map. The CESAGAN agent, by contrast, learned the distribution of different tiles from the entire map.

\section{Discussion}
%image inpainting, and stuff

The core underlying idea of the PCG method described here is that generation can be seen as repair. The trained generator is faced with a sequence of broken level segments, and asked to determine the edit that would best repair the broken segment. This is why the levels first need to be ``destroyed'' in order to train the model: as we already know the correct form of each segment, we can create a dataset of repair actions to learn. Given that any map can be destroyed in a very large number of different ways, one can view the process of dataset generation as a form of data augmentation. This helps to explain the strong results we've seen from generation based training on a very small number of levels. While we have not seen the idea of creation as repair expressively stated in computational creativity or creative AI research, it relates to many other ideas. For example, many schools of art teaching focus on how to gradually improve an artifact from an imperfect state.

It is interesting to compare the Path of Destruction method with image inpainting techniques. For centuries, conservators have studied how to best restore art pieces by inpainting parts that are missing or destroyed due to e.g. deterioration or vandalism. In recent years, a burgeoning subfield of AI poses inpainting as a computational problem and advances machine learning-based methods to solve it~\cite{bertalmio2000image}. These methods typically take a set of images and create training data by randomly removing parts of the images. The model is thus tasked with regenerating the missing parts. In that sense, the overall procedure of image inpainting is similar to the Path of Destruction method presented here. However, in our method, each edit action pertains only to a single tile (or pixel) based on a limited neighboring field of view.

Diffusion Models are another obvious point of comparison\footnote{The Path of Destruction technique was developed independently from Diffusion Models, and the similarity was only pointed out to the authors after a first preprint of this paper had been posted.}. Those methods are similar to Path of Destruction in that an agent learns a task through repair steps generated from destroyed images~\cite{Dhariwal2021}. However, there are some differences between PoD and diffusion models. Perhaps the most significant difference is that diffusion models add Gaussian noise to portion of the image in one step. By contrast, in PoD, the level is destroyed iteratively, one tile at a time. Moreover, PoD destroys the level of each episode towards a goal whereas, with diffusion models, the image is destroyed towards a distribution. While diffusion models could be trained on levels instead of images, they normally require extremely large datasets to generalize. Achieving diffusion model generalization using a small dataset of levels constitutes an interesting area of future exploration.

We can also contrast our novel method with PCG methods based on predicting a tile based on immediate neighbor tiles. This includes the widely-used Wave Function Collapse (WFC) algorithm~\footnote{\url{https://github.com/mxgmn/WaveFunctionCollapse}}, a constraint-solving algorithm that learns local constraints from a small amount of data and uses them to create new content that recreates local patterns stochastically~\cite{karth2017wavefunctioncollapse}. It also includes methods based on cellular automata. Specifically, a variation known as Neural Cellular Automata (NCA) can be trained using gradient descent to recreate images~\cite{mordvintsev2020growing}. It has also recently been shown that evolutionary methods can be used to create NCAs that create playable levels~\cite{earle2022illuminating}. Yet another related method is Multidimensional Markov Chains (MdMC), which directly learn the conditional probabilities of neighboring tiles~\cite{snodgrass2016controllable}. Compared to these methods, the Path of Destruction takes a larger part of the level into account as part of the input observation. We believe the data augmentation performed by the Path of Destruction method leads to better few-shot generalization compared to methods which learn directly from the source levels.

\section{Conclusion}
We have presented the Path of Destruction, a procedural content generation method which can train iterative level generators from existing levels. The key innovation is the data generation (or augmentation) method, which makes a large number of small changes to destroy a level and creates a dataset of the destruction played backwards, mapping from a destroyed level to a single repair action. By training convolutional networks on the generated data, we created level generators that reliably produce playable and unique levels for three different 2D games. It is notable that PoD generators can generate a wide variety of levels even when trained on a limited number of goal levels. This was most evident when we measured the PoD agents' performance against the CESAGAN baseline. The results illustrated that the PoD agents significantly outperformed the benchmark in terms of the percentage of unique playable maps generated.

A number of developments of the method proposed here are possible and should be investigated. It is worth further exploring different neural architectures, observation sizes, and representation types. One could also envision using evolutionary search or quality diversity to produce a diverse archive of training episodes to increase the uniqueness of the generated levels.

\bibliographystyle{IEEEtran}
\bibliography{IEEEfull,bibliography}

% Generated by IEEEtran.bst, version: 1.12 (2007/01/11)
\begin{thebibliography}{10}
\providecommand{\url}[1]{#1}
\csname url@samestyle\endcsname
\providecommand{\newblock}{\relax}
\providecommand{\bibinfo}[2]{#2}
\providecommand{\BIBentrySTDinterwordspacing}{\spaceskip=0pt\relax}
\providecommand{\BIBentryALTinterwordstretchfactor}{4}
\providecommand{\BIBentryALTinterwordspacing}{\spaceskip=\fontdimen2\font plus
\BIBentryALTinterwordstretchfactor\fontdimen3\font minus
  \fontdimen4\font\relax}
\providecommand{\BIBforeignlanguage}[2]{{%
\expandafter\ifx\csname l@#1\endcsname\relax
\typeout{** WARNING: IEEEtran.bst: No hyphenation pattern has been}%
\typeout{** loaded for the language `#1'. Using the pattern for}%
\typeout{** the default language instead.}%
\else
\language=\csname l@#1\endcsname
\fi
#2}}
\providecommand{\BIBdecl}{\relax}
\BIBdecl

\bibitem{summerville2018procedural}
A.~Summerville, S.~Snodgrass, M.~Guzdial, C.~Holmg{\aa}rd, A.~K. Hoover,
  A.~Isaksen, A.~Nealen, and J.~Togelius, ``Procedural content generation via
  machine learning (pcgml),'' \emph{IEEE Transactions on Games}, vol.~10,
  no.~3, pp. 257--270, 2018.

\bibitem{shaker2016constructive}
N.~Shaker, A.~Liapis, J.~Togelius, R.~Lopes, and R.~Bidarra, ``Constructive
  generation methods for dungeons and levels,'' in \emph{Procedural Content
  Generation in Games}.\hskip 1em plus 0.5em minus 0.4em\relax Springer, 2016,
  pp. 31--55.

\bibitem{togelius2011search}
J.~Togelius, G.~N. Yannakakis, K.~O. Stanley, and C.~Browne, ``Search-based
  procedural content generation: A taxonomy and survey,'' \emph{IEEE
  Transactions on Computational Intelligence and AI in Games}, vol.~3, no.~3,
  pp. 172--186, 2011.

\bibitem{smith2011answer}
A.~M. Smith and M.~Mateas, ``Answer set programming for procedural content
  generation: A design space approach,'' \emph{IEEE Transactions on
  Computational Intelligence and AI in Games}, vol.~3, no.~3, pp. 187--200,
  2011.

\bibitem{jain2016autoencoders}
R.~Jain, A.~Isaksen, C.~Holmg{\aa}rd, and J.~Togelius, ``Autoencoders for level
  generation, repair, and recognition,'' in \emph{Proceedings of the ICCC
  workshop on computational creativity and games}, 2016, p.~9.

\bibitem{volz2018evolving}
V.~Volz, J.~Schrum, J.~Liu, S.~M. Lucas, A.~Smith, and S.~Risi, ``Evolving
  mario levels in the latent space of a deep convolutional generative
  adversarial network,'' in \emph{Proceedings of the genetic and evolutionary
  computation conference}, 2018, pp. 221--228.

\bibitem{dahlskog2014linear}
S.~Dahlskog, J.~Togelius, and M.~J. Nelson, ``Linear levels through n-grams,''
  in \emph{Proceedings of the 18th International Academic MindTrek Conference:
  Media Business, Management, Content \& Services}, 2014, pp. 200--206.

\bibitem{summerville2016super}
A.~J. Summerville and M.~Mateas, ``Super mario as a string: Platformer level
  generation via lstms,'' in \emph{Proceedings of DiGRA/FDG}, 2016.

\bibitem{khalifa2020pcgrl}
A.~Khalifa, P.~Bontrager, S.~Earle, and J.~Togelius, ``Pcgrl: Procedural
  content generation via reinforcement learning,'' in \emph{Proceedings of the
  AAAI Conference on Artificial Intelligence and Interactive Digital
  Entertainment}, vol.~16, 2020, pp. 95--101.

\bibitem{earle2021learning}
S.~Earle, M.~Edwards, A.~Khalifa, P.~Bontrager, and J.~Togelius, ``Learning
  controllable content generators,'' in \emph{2021 IEEE Conference on Games
  (CoG)}.\hskip 1em plus 0.5em minus 0.4em\relax IEEE, 2021, pp. 1--9.

\bibitem{delarosa2021mixed}
O.~Delarosa, H.~Dong, M.~Ruan, A.~Khalifa, and J.~Togelius, ``Mixed-initiative
  level design with rl brush,'' in \emph{International Conference on
  Computational Intelligence in Music, Sound, Art and Design (Part of
  EvoStar)}.\hskip 1em plus 0.5em minus 0.4em\relax Springer, 2021, pp.
  412--426.

\bibitem{ye2020rotation}
C.~Ye, A.~Khalifa, P.~Bontrager, and J.~Togelius, ``Rotation, translation, and
  cropping for zero-shot generalization,'' in \emph{2020 IEEE Conference on
  Games (CoG)}.\hskip 1em plus 0.5em minus 0.4em\relax IEEE, 2020, pp. 57--64.

\bibitem{earle2022illuminating}
S.~Earle, J.~Snider, M.~C. Fontaine, S.~Nikolaidis, and J.~Togelius,
  ``Illuminating diverse neural cellular automata for level generation,'' in
  \emph{Proceedings of the Genetic and Evolutionary Computation Conference},
  2022, pp. 68--76.

\bibitem{torrado2020bootstrapping}
R.~R. Torrado, A.~Khalifa, M.~C. Green, N.~Justesen, S.~Risi, and J.~Togelius,
  ``Bootstrapping conditional gans for video game level generation,'' in
  \emph{2020 IEEE Conference on Games (CoG)}.\hskip 1em plus 0.5em minus
  0.4em\relax IEEE, 2020, pp. 41--48.

\bibitem{snodgrass2017procedural}
S.~Snodgrass and S.~Ontan{\'o}n, ``Procedural level generation using
  multi-layer level representations with mdmcs,'' in \emph{2017 IEEE conference
  on computational intelligence and games (CIG)}.\hskip 1em plus 0.5em minus
  0.4em\relax IEEE, 2017, pp. 280--287.

\bibitem{bertalmio2000image}
M.~Bertalmio, G.~Sapiro, V.~Caselles, and C.~Ballester, ``Image inpainting,''
  in \emph{Proceedings of the 27th annual conference on Computer graphics and
  interactive techniques}, 2000, pp. 417--424.

\bibitem{Dhariwal2021}
\BIBentryALTinterwordspacing
P.~Dhariwal and A.~Nichol, ``Diffusion models beat gans on image synthesis,''
  \emph{CoRR}, vol. abs/2105.05233, 2021. [Online]. Available:
  \url{https://arxiv.org/abs/2105.05233}
\BIBentrySTDinterwordspacing

\bibitem{karth2017wavefunctioncollapse}
I.~Karth and A.~M. Smith, ``Wavefunctioncollapse is constraint solving in the
  wild,'' in \emph{Proceedings of the 12th International Conference on the
  Foundations of Digital Games}, 2017, pp. 1--10.

\bibitem{mordvintsev2020growing}
A.~Mordvintsev, E.~Randazzo, E.~Niklasson, and M.~Levin, ``Growing neural
  cellular automata,'' \emph{Distill}, vol.~5, no.~2, p. e23, 2020.

\bibitem{snodgrass2016controllable}
S.~Snodgrass and S.~Ontan{\'o}n, ``Controllable procedural content generation
  via constrained multi-dimensional markov chain sampling.'' in \emph{IJCAI},
  2016, pp. 780--786.

\end{thebibliography}

\end{document}